\documentclass{article}
\usepackage{iclr2027_conference,times}
\usepackage{lineno}
\usepackage{graphicx}
\usepackage{booktabs}
\usepackage{multirow}
\usepackage{amsmath,amssymb}
\usepackage{microtype}
\usepackage{url}
\usepackage{xspace}
\usepackage{enumitem}
\usepackage{caption}
\usepackage{xcolor}
\usepackage[colorlinks=true,allcolors=blue!55!black]{hyperref}

\graphicspath{{figures/}}


\newcommand{\zL}{Z_L}
\newcommand{\zH}{Z_H}
\newcommand{\Early}{\textsc{Early}\xspace}
\newcommand{\Late}{\textsc{Late}\xspace}
\newcommand{\Persistent}{\textsc{Persistent}\xspace}
\newcommand{\gnat}{\gamma^{\mathrm{nat}}}
\newcommand{\dout}{d^{\mathrm{out}}}
\newcommand{\ci}[2]{{\small$[#1,\,#2]$}}
\newcommand{\cip}[2]{{\small[#1,\,#2]}}

\newcommand{\Matched}{\textsc{Matched}\xspace}
\newcommand{\Unmatched}{\textsc{Unmatched}\xspace}
\newcommand{\Unsolved}{\textsc{Unsolved}\xspace}

\title{When Recursive Models Finish Computing}

\author{
Hare Krishna$^{1}$,
Shubham Singh$^{2}$,
Stephen Ebert$^{3}$,
Hao-Yu Sun$^{4,5}$\\[2mm]
$^{1}$Weinberg Institute, Department of Physics, University of Texas at Austin\\
$^{2}$Department of Urban and Environmental Policy Planning, Tufts University\\
$^{3}$Zyphra Technologies, San Francisco, CA, USA\\
$^{4}$Mathematics Department, Austin Community College\\
$^{5}$SpaceXAI, USA\\
[1mm]
{\scriptsize
\texttt{hkrishna.phy@gmail.com, shubhams12101@gmail.com, stephenebert@gmail.com, hkdavidsun@utexas.edu}
}
}

\iclrfinalcopy

\begin{document}

\maketitle
\fancyhead{}
\renewcommand{\headrulewidth}{0pt}

\begin{abstract}
Recursive models can continue updating their latent states beyond their nominal inference budget, so an incorrect output at that budget does not show whether computation is unfinished or has entered a persistently unsuccessful regime. We study the dynamics of completion in attention- and MLP-based Tiny Recursive Models (TRMs) on 1,000 hard Sudoku puzzles. Extending recurrence from the nominal 16 steps to 512 steps increases cumulative exact-solve accuracy from $59.2\%$ to $87.5\%$ for the attention model and from $74.4\%$ to $91.9\%$ for the MLP model, solving more than two-thirds of the puzzles unsolved in the nominal budget. Across both architectures, latent-state motion drops sharply after the first exact solution. Completed states are typically locally contractive along the trajectory direction, even though the same local Jacobian retains strongly expanding directions. We characterize this phenomenon as \emph{trajectory-conditioned anisotropic stability}. Perturbation experiments confirm this directional stability across both models. The multi-step fate of the maximally expanding direction differs: it is absorbed within 16 steps in the attention model but persists longer in the MLP model. The anisotropic-stability pattern also holds for a second attention checkpoint. Together, these results distinguish nominal-budget failure from completed computation and identify a common dynamical signature of completion across two recurrent architectures.



\end{abstract}

\section{Introduction}
Recursive models compute by iteratively updating a latent state, applying the same learned transition function at each step. This provides additional computational depth through weight sharing \citep{dehghani2019universal,giannou2023looped} and enables computation to continue across multiple iterations \citep{graves2016adaptive,schwarzschild2021learn,bansal2022endtoend,geiping2025recurrentdepth}. Hierarchical Reasoning Models (HRMs) and Tiny Recursive Models (TRMs) use this principle to build compact reasoning systems \citep{wang2025hrm,jolicoeurmartineau2025trm}. In such models, the number of recurrent steps determines how long the model can compute.

This creates a fundamental ambiguity. An incorrect answer at the end of the inference budget can mean two different things. The model may have failed, or it may simply need more computation \citep{anil2022path}. Conversely, a correct output does not tell us whether the underlying state has settled. The model may already have reached the answer, while its internal state continues to change \citep{knutson2026logical}. Endpoint accuracy alone does not distinguish \emph{failure}, \emph{completion}, or \emph{stability}.

Recurrent computation has long been studied via its state-space dynamics. Fixed points, attractors, and local Jacobians have been used to understand how recurrent models represent and transform information \citep{sussillo2013opening,maheswaranathan2019line}. Equilibrium models make the connection between computation and fixed points explicit \citep{bai2019deepequilibriummodels,huang2026equilibrium}. Recent mechanistic studies of HRMs and TRMs have identified attractor structure, failure modes, spurious fixed points, and abrupt changes in solution quality \citep{balwani2026attractor,ren2026reasoning}. Complementary work on recurrent-depth reasoners has studied settling and the conditions under which additional test-time depth remains useful and stable \citep{viakhirev2026thinkshallowsolvedeep}, while recent dynamical-systems analyses connect prolonged reasoning to transient chaos and fractal basin structure \citep{lai2026fractal}.


We ask a complementary question: \emph{what changes in the recurrent state when a particular computation finishes?} Rather than comparing all trajectories at the same recurrent step, we align each problem to the step at which it is first solved. We then study completion, latent motion, and local stability separately. Completion is the first step at which the exact solution is produced. Latent motion measures how much the recurrent latent state changes from one step to the next. Stability describes how perturbations of that state grow or shrink. We measure it using the local Jacobian and controlled finite perturbations. This lets us distinguish directional from uniform stability. The model may become stable along the direction its own computation follows, even though the local Jacobian contains an expanding direction.

We study this question in attention and MLP-based TRMs on the same 1,000 hard Sudoku puzzles. Both models have a nominal budget of 16 recurrent steps. We continue the unchanged recurrence to 512 steps. Cumulative exact solution rises from $59.2\%$ to $87.5\%$ for the attention model and from $74.4\%$ to $91.9\%$ for the MLP model. Among the puzzles that are unresolved at step 16, $69.4\%$ and $68.4\%$ are solved later, respectively. In both models, more than two-thirds of the apparent failures in the nominal budget are unfinished computations. This separates the puzzles into three completion groups. \Early solvers are solved by step 16. \Late solvers are solved between steps 17 and 512. \Persistent puzzles remain unsolved through step 512. 
Once aligned with their own solution time, \Early and \Late solvers show a similar completion pattern. Their latent motion decreases, and they enter a directionally stable regime. The \Persistent trajectories remain much more active and show a different stability profile.
\paragraph{Completion has a directional stability signature.}
Latent-state motion drops sharply after completion. Completed states are locally contractive along the direction in which the model is moving. However, the same local Jacobian retains strongly expanding directions. The recurrence, therefore, does not become uniformly contractive when the task is solved. Perturbation experiments confirm this directional contrast. Perturbations along the most expanding direction grow much more than perturbations of the same size along the model's natural direction of motion. We call this coexistence \emph{trajectory-conditioned anisotropic stability}. The computation becomes stable along the direction of motion while the recurrent map retains locally expanding directions.

We repeat the full analysis in the attention and MLP models, using the same puzzles, recurrent schedule, probe protocol, and statistical procedures. The qualitative features replicate: unsolved trajectories remain active, completed trajectories settle, the natural direction is locally contractive after completion, and expanding directions remain in the Jacobian. But 
after completion, perturbations along the most expanding direction are absorbed within 16 steps in the attention model (\S\ref{sec:geometry}) and persist in the MLP model (\S\ref{sec:mlp}).

We also present supporting evidence beyond the two primary hard Sudoku runs. A second attention checkpoint, Attention-B, shows a similar anisotropic-stability pattern (Appendix~\ref{app:architecture_summary}, Table~\ref{tab:cross}). In Easy Sudoku, most puzzles are solved within the first few recurrent steps and fall into the \Early-solver regime. Their latent dynamics closely resemble those of the \Early solvers in Hard Sudoku (Appendix~\ref{app:easy}). In Maze-Hard, a separate $30\times30$ grid task, most mazes are also solved within the first few recurrent steps and therefore behave mainly like \Early solvers. Maze-Hard also shows that exact-match accuracy can be misleading. Some predictions differ from the ground-truth reference path, but are still valid solutions (Appendix~\ref{app:maze}). The full extended-recurrence, Jacobian, and perturbation analyses are therefore centered on Hard Sudoku, where both early and late completion are well represented. 


\section{Experimental Framework}
\subsection{Models, task, and extended recurrence}
\label{sec:setup}
We use the TRM architecture of \citet{jolicoeurmartineau2025trm} (see its Figures 1 and 3) and evaluate two trained checkpoints on the 1,000 hard Sudoku puzzles. \mbox{\textbf{Attention-A}} is a $6.83$M-parameter recurrent transformer checkpoint \mbox{\citep{sanjin2024checkpoint,trmrepository}}. The \mbox{\textbf{MLP}} checkpoint replaces attention with sequence-axis MLP mixing and removes positional encoding \mbox{\citep{sanjin2025checkpointmlp}}. Both use the same recurrent schedule, probe protocol, and statistics. The configuration details are shown in Appendices~\mbox{\ref{app:setup}} and~\mbox{\ref{app:mlp_setup}}.

The 1{,}000 puzzles come from the Sudoku-Extreme test split \citep{wang2025hrm}. All belong to its \mbox{\texttt{puzzles4\_forum\_hardest\_1905}} source, the \mbox{\texttt{forum\_hardest\_1905}} list collected on the Enjoy Sudoku players' forum and distributed with the \mbox{\texttt{tdoku}} benchmark suite \citep{tdoku}. Both checkpoints have a nominal 16-step outer budget. We reproduce the 16-step evaluation and continue the learned recurrence to a diagnostic horizon of 512 steps with the same weights, inputs, and pre-processing. 


\subsection{Recurrent state, motion, and completion groups}

In the recurrent step $t$, the model carries two latent states, $\zH^{(t)}$ and $\zL^{(t)}$. We combine them into a single state,
\begin{equation}
s_t=(\zH^{(t)},\zL^{(t)}),
\qquad
s_{t+1}=F_\theta(s_t;x).
\end{equation}
All state-space analyses use this joint state, with dimension
$2\times97\times512$. We measure how much each latent state component $\zH,\zL$ changes between consecutive steps using the relative update
\begin{equation}
\delta_B(t)=
\frac{\|B^{(t)}-B^{(t-1)}\|_2}
{\|B^{(t-1)}\|_2},
\qquad
B\in\{\zH,\zL\}.
\end{equation}
A small value of $\delta_B(t)$ means that the latent state changes little from one step to the next. For each puzzle $i$, let $\tau_i$ denote the first recurrent step at which the decoded Sudoku is exactly correct. We divide the puzzles into three groups: \Early solvers have $\tau_i\le16$, \Late solvers have $16<\tau_i\le512$, and \Persistent puzzles remain unsolved through step 512. When comparing trajectories around completion, we measure time relative to the first solve: $r = t - \tau_i$. Here, $r=0$ is the solving step; negative values precede completion, and positive values follow.

\subsection{Local stability along the trajectory}
To characterize the dynamics along the model's trajectory, we linearize the recurrent map at each visited state,
\begin{equation}
J_t
=
\left.
\frac{\partial F_\theta}{\partial s}
\right|_{s_t}.
\end{equation}

We use it to distinguish stability along the direction followed by the model's trajectory from the largest amplification available in the surrounding state space. We define the unit trajectory direction
\begin{equation}
    \dout_t
    =
    \frac{s_{t+1}-s_t}
    {\|s_{t+1}-s_t\|_2},
\end{equation}
whenever $\|s_{t+1}-s_t\|_2$ is numerically nonzero. Its one-step Jacobian gain is 
\begin{equation}
    \gnat_t
    =
    \|J_t \dout_t\|_2.
\end{equation}
Perturbations along the trajectory direction therefore contract to first order when $\gnat_t<1$ and grow when $\gnat_t>1$. Since $\Delta_{t+1}\approx J_t\Delta_t$ for $\Delta_t=s_{t+1}-s_t$, $\gnat_t$ approximates the ratio $\|\Delta_{t+1}\|_2/\|\Delta_t\|_2$ of successive updates wherever the linearization is accurate. 
Thus, $\gnat_t$ provides a local dynamical interpretation of shrinking or growing latent updates. Its additional value comes from comparing the gain along the direction actually followed by the trajectory with the amplification available along other directions at the same state. In particular, $\gnat_t<1$ indicates contraction only along the local trajectory direction and does not imply contraction of the full surrounding state space.

To measure the strongest local amplification available at the same state, we compute the spectral norm of the Jacobian,
\begin{equation}
    \sigma_{\max}(J_t)
    =
    \max_{\|v\|_2=1}
    \|J_t v\|_2
    =
    \sigma_1(t).
\end{equation}
Let $v_1(t)$ denote a corresponding leading right singular vector,
\begin{equation}
    v_1(t)
    \in
    \arg\max_{\|v\|_2=1}
    \|J_t v\|_2.
\end{equation}
Under one linearized recurrent step,
\begin{equation}
    J_t v_1(t)
    =
    \sigma_1(t)\,u_1(t),
\end{equation}
where $u_1(t)$ is the corresponding leading left singular vector. Here $v_1(t)$ is the most expanding local perturbation direction, $\sigma_1(t)$ is its one-step amplification factor, and $u_1(t)$ is its output direction.

The comparison between $\gnat_t$ and $\sigma_1(t)$ distinguishes stability along the trajectory from worst-case local stability. In particular,
\begin{equation}
    \gnat_t < 1
    \qquad \text{and} \qquad
    \sigma_1(t) > 1
\end{equation}
indicates anisotropic local dynamics: to first order, the trajectory-aligned direction is contracting even though expanding directions remain available in the surrounding state space.

We compute the leading singular triplet $(\sigma_1,v_1,u_1)$ without explicitly constructing the Jacobian. We apply matrix-free power iteration to $J_t^\top J_t$ using forward-mode Jacobian-vector products and reverse-mode vector-Jacobian products. Implementation details, convergence checks, and numerical validation appear in Appendix~\ref{app:jacobian}. We also compute the same directional gains for probes restricted to either the $\zH$ or $\zL$ block.

The squared alignment $A_1(t)=|\langle \dout_t,v_1(t)\rangle|^2$ measures how closely the model's own trajectory aligns with the most expanding direction.  We compare this $A_1(t)$ with an isotropic random-direction baseline evaluated at the same state.

Finally, we measure whether the amplified direction persists across recurrent steps. We define
\begin{equation}
T_1(t)=|\langle u_1(t),v_1(t+1)\rangle|^2.
\end{equation}
A large $T_1$ means that the direction produced by maximal amplification at step $t$ is aligned with the most expanding input direction at the next step. A small value means that this direction is largely rotated away.
To probe stability beyond a single step, we measure finite-horizon growth by perturbing $s_t$ along a unit direction $v$, with $\tilde s_t=s_t+\varepsilon v$. After evolving both trajectories for $k$ recurrent steps, let
\begin{equation}
\delta s_{t+k}^{(v)}=\tilde s_{t+k}-s_{t+k}\nonumber
\end{equation}
denote their separation. We define
\begin{equation}
\lambda_k(v)=
\frac{1}{k}
\log\left(
\frac{\|\delta s_{t+k}^{(v)}\|_2}{\varepsilon}
\right).
\end{equation}
Negative values indicate decay of the perturbation over $k$ steps, while positive values indicate growth. This is a finite-horizon, trajectory-dependent measure of perturbation growth.


\begin{figure}[t]
\centering
\includegraphics[width=\textwidth]{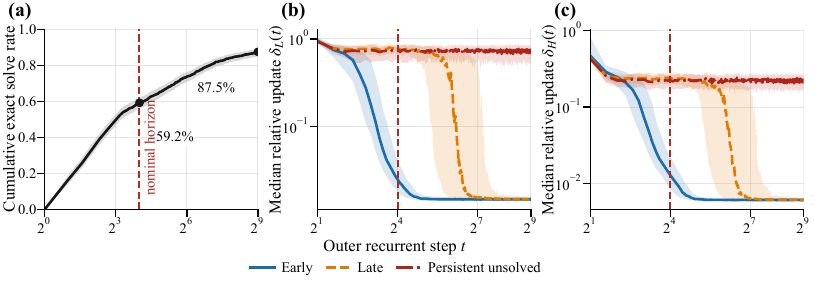}
\caption{\textbf{The nominal horizon truncates active computation.}
(a)~Cumulative exact solve rate over 512 outer steps with Wilson bands; the dashed line is the nominal 16-step budget. (b,c)~Median relative latent update with interquartile bands for the three completion groups. \Late solvers are as mobile as \Persistent ones at step 16 and settle only when they solve. \Persistent trajectories never settle.}
\label{fig:extended}
\end{figure}

\section{Results}
\subsection{Extended recurrence reveals unfinished computation}
\label{sec:extended}

Continuing the same recurrence beyond its nominal 16-step horizon changes the picture substantially. The exact solve rate rises from 59.2\% at step 16 to 87.5\% by step 512 (Figure~\ref{fig:extended}a). Of the 408 puzzles that are still unsolved at step 16, 283 are solved later. About two-thirds of the apparent failures at the nominal horizon are not failures at all. They are computations that have not yet finished. \Late solvers typically need many more steps: their median first solve occurs at step 62 (interquartile 33-130), although some solve much later. 

The latent dynamics of these late solvers support the same interpretation. At step 16 (Figure~\ref{fig:extended}b,c), late solvers are still moving strongly and look much more like persistently unsolved trajectories than early solvers. For example, their median $\delta_L$ is $0.763$, compared with only $0.025$ for early solvers. By step 512, however, late solvers have settled to the same low-motion regime as early solvers, while persistent trajectories remain highly active (Figure~\ref{fig:extended}b,c). This yields three populations to analyze: \Early solvers, \Late solvers, and \Persistent puzzles.

Puzzle difficulty does not explain this separation. Persistent puzzles are not systematically harder according to the dataset ratings. If anything, their median rating is slightly lower. Full difficulty statistics are reported in Appendix~\ref{app:difficulty}.

\begin{figure}[t]
\centering
\includegraphics[width=\textwidth]{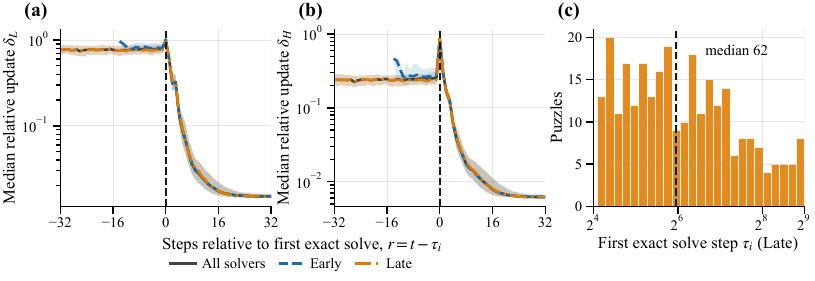}
\caption{\textbf{Settling is aligned to each puzzle's own completion.}
(a,b)~Median relative update on the solve-aligned axis $r=t-\tau_i$, with interquartile bands. Motion is roughly constant before completion, spikes at $r=0$, and then falls by more than an order of magnitude. The \Early and \Late curves coincide once aligned, although their absolute solve times differ by an order of magnitude. (c)~Distribution of $\tau_i$ for \Late solvers.
}
\label{fig:aligned}
\end{figure}
\subsection{Latent settling aligns with computational completion}
\label{sec:settling}
When trajectories are aligned to each puzzle's first solve time $\tau_i$, the \Early and \Late solvers follow a similar pattern (Figure~\ref{fig:aligned}). For \Late solvers, latent motion stays roughly constant before completion and then drops sharply after the solution is reached. In $\zL$, the median motion after completion is less than half of the pre-solve level. After completion, \Late solvers settle to the same low-motion regime as \Early solvers.  

\paragraph{The solving step is a large reorganization.}
The strongest change occurs exactly at the solving step, $r=0$ (Figure~\ref{fig:aligned}a,b). For \Late solvers, the median $\delta_H$ rises from $0.238$ at $r=-1$ to $0.878$ at $r=0$. For $99.3\%$ of \Late solvers, the $\zH$ update at the solving step is larger than at every earlier step in the trajectory. The effect is much weaker in $\zL$, where the solving step is the largest update for only $35\%$ of puzzles. This spike is not explained simply by a change in the decoded grid.  Many earlier steps also change the prediction without solving the puzzle, but their $\zH$ updates are much smaller. 
The solving step, therefore, marks an unusually large reorganization of the high-level latent state. It is followed immediately by a sharp reduction in motion.

\paragraph{Solutions are usually stable.}
Solved puzzles stay in the recurrence, and almost all of them remain solved. Of the 875 puzzles that solve at least once, 859 never lose the solution again. The remaining 16 temporarily revert, but all recover by step 512. We therefore use the first exact solution as the completion time.
\subsection{Completed computation is directionally stable, not uniformly
contractive}
\label{sec:stability}
Unlike our earlier analysis of latent settling, Jacobian analysis asks whether the perturbations along different state-space directions are amplified or contracted. The completed trajectories are typically locally contracting along their own direction of motion, while the same local Jacobian has strongly expanding directions. We evaluate the Jacobian at 750 visited states, corresponding to 150 puzzles at five fixed recurrent steps each. The recovered singular vectors have unit norm up to numerical precision. Additional numerical checks and implementation details are in Appendix~\ref{app:jacobian}.


\begin{table}[t]
\centering\small
\setlength{\tabcolsep}{3.6pt}
\caption{Local stability at the nominal horizon and at step 512, by completion
group. $\gnat$ is the Jacobian gain along the model's own direction of travel.  $\sigma_{\max}$ is the worst-case gain over all directions in the same state. $\Pr[\gnat<1]$ represents the fraction of states in that group having $\gnat<1$.  }
\label{tab:stability}
\begin{tabular}{@{}llccc@{}}
\toprule
Step & Group & $\gnat$ & $\Pr[\gnat<1]$ & $\sigma_{\max}$ \\
\midrule
\multirow{3}{*}{16}
 & \Early      & 0.83  & 0.74 & \phantom{0}25.7 \\
 & \Late       & 2.28  & 0.24 & 392.2 \\
 & \Persistent & 3.44  & 0.12 & 529.4 \\
\midrule
\multirow{3}{*}{512}
 & \Early      & 0.25  & 0.96 & \phantom{0}26.8 \\
 & \Late       & 0.31  & 0.92 & \phantom{0}26.9 \\
 & \Persistent & 2.09  & 0.30 & 316.9 \\
\bottomrule
\end{tabular}
\end{table}

\begin{figure}[t]
\centering
\includegraphics[width=\textwidth]{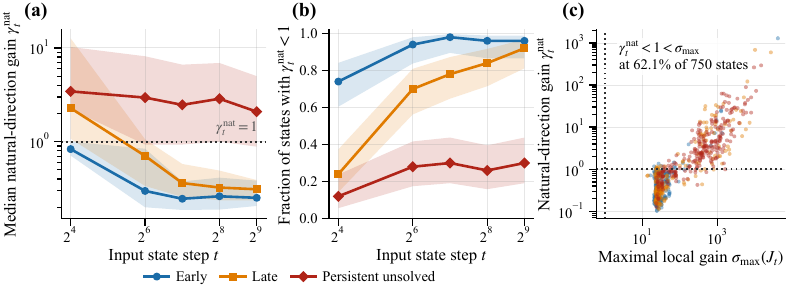}
\caption{\textbf{Stability is conditioned on the direction of travel.}
(a)~Median natural-direction gain $\gnat_t$ by group and input-state step, with interquartile bands. The dotted line is $\gnat=1$. (b)~Fraction of states with $\gnat_t<1$, with Wilson bands. (c)~Every probed state: $\gnat_t$ against
$\sigma_{\max}(J_t)$. No state has $\sigma_{\max}<1$, yet most completed states
sit below $\gnat=1$.}
\label{fig:stability}
\end{figure}

Table~\ref{tab:stability} and Figure~\ref{fig:stability} show the main result. At step 16, the three completion groups have different natural-direction gains. \Early solvers have a median $\gnat$ of $0.83$, whereas \Late and \Persistent trajectories remain locally expansive, with medians of $2.28$ and $3.44$.

\Late solvers change once they finish computing. By step 512 (Table~\ref{tab:stability}), their median $\gnat$ has fallen to $0.31$, close to the \Early value of $0.25$ (Figure~\ref{fig:stability}a). Most completed states then contract locally in the direction the model is moving.  $96\%$ of \Early states and $92\%$ of \Late states have $\gnat<1$.  
At that point, every \Late solver has completed, so the two groups occupy the same directionally stable regime. Because Figure~\ref{fig:stability} compares fixed recurrent steps, it shows where the groups end up rather than when the change happens. The solve-aligned transition is in latent motion: aligned to their own solve step, \Late and \Early solvers show the same drop in latent updates (Figure~\ref{fig:aligned}a,b). \Persistent trajectories remain different, with median $\gnat=2.09$.

This stability is not a property of the full recurrent map. The worst-case Jacobian gain remains larger than one at every probed state. At step 512, $\sigma_{\max}$ is about $27$ for both completed groups and above $300$ for \Persistent trajectories. Contraction along the natural direction and expansion along some other direction coexist at $62.1\%$ of probed states (Figure~\ref{fig:stability}c). 

Probes restricted to $\zH$ or $\zL$ show the same pattern. At step 512, the median $\zL$-probe gain is $0.023$ for \Early solvers and $0.021$ for \Late solvers, compared with $0.309$ for \Persistent trajectories. Finite-horizon growth over the 16 steps after step 256 is also negative for both completed groups and positive for \Persistent trajectories. Full results are reported in Appendix~\ref{app:blocks}.


\begin{figure}[t]
\centering
\includegraphics[width=\textwidth]{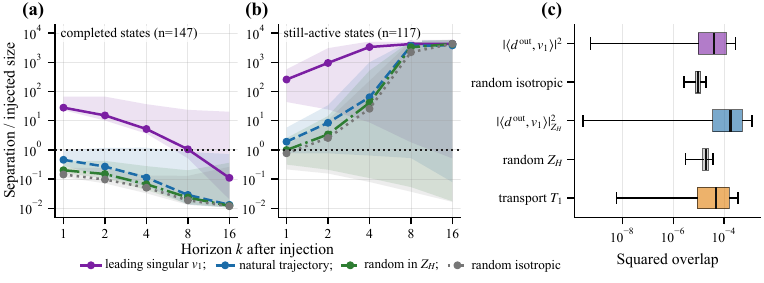}
\caption{\textbf{An expanding direction is present, but the  trajectory is not orthogonal to it.} Norm-matched perturbations ($10^{-4}$ of the state norm) along four directions, followed for $k$ further steps, at (a) states whose computation has completed and (b) states still computing. In (b) every direction, including a random one, reaches the scale of the state itself. These curves measure trajectory divergence. (c) Squared overlaps: the trajectory's overlap with $v_1$ is tiny but distinctly above the isotropic baseline. The transport of the leading direction across one step is of the same order. The ``random isotropic'' is the overlap of $v_1$ with a random unit direction drawn uniformly from the full hidden-state space, and ``random \(\zH\)'' uses a random unit direction drawn from the \(\zH\) subspace only.}
\label{fig:geometry}
\end{figure}

\subsection{Fate and geometry of the expanding direction}
\label{sec:geometry}
Although $\sigma_{\max}>1$ at every probed state, this strong local expansion does not destabilize completed trajectories. We measure how the trajectory aligns with the most-expanding direction $v_1$ and track perturbations injected along that direction. 

\paragraph{The trajectory does not avoid $v_1$.}
The alignment $A_1$ is small, with median $4.15\times10^{-5}$ over 750 states. However, this is about four times larger than the isotropic random-direction baseline of $9.63\times10^{-6}$. The trajectory is therefore not orthogonal to the direction of expansion. The overlap is small because the state space has $\sim\!10^5$ dimensions.

\paragraph{A $v_1$ perturbation and its later fate depend on computation completeness.} A perturbation of size $10^{-4}$ of the state norm along $v_1$ grows by $44.4\times$ after one step, against $0.75\times$ along the natural direction (medians over the 180 fixed-step probe states). Paired state by state over all 264 probes, the median $v_1$-to-natural ratio is about $83\times$. The median ratio of one-step $v_1$ amplification to the estimated $\sigma_{\max}$ is 1.00.
Over the 147 completed probe states (99 fixed-step and 48 solve-aligned), the median $v_1$ perturbation grows by about $28\times$ initially but falls to $0.11$ of its injected size by $k=16$. At states that are still computing, perturbations in all tested directions eventually produce large trajectory separation (Figure~\ref{fig:geometry}a,b).

\paragraph{The expanding direction is not carried forward.}
The amplified output direction $u_1(t)$ has very little overlap with the next step's most expanding input direction.  The transport defined earlier is $T_1(t)\approx 4.9\times10^{-5}$. Thus, a direction that expands strongly in one step is largely rotated away before the next step. Strong one-step expansion, therefore, need not produce sustained multi-step growth. Non-normal recurrent dynamics can amplify a perturbation transiently without sustained growth \citet{kerg2019nnrnn}, although we do not test whether this mechanism operates here. The leading input direction $v_1$ lies almost entirely in $\zH$, while its amplified output lies mostly in $\zL$. The strongest local amplification, therefore, acts mainly from $\zH$ into $\zL$. This is also consistent with the latent supported probes in
Appendix~\ref{app:blocks}.
\subsection{Cross-architecture replication in an MLP recurrent model}
\label{sec:mlp}
We repeat the full analysis on the MLP model with the same task, data, completion definitions, diagnostic horizon, and statistics (\S\ref{sec:setup}). Only the model and checkpoints change. 
 
\begin{figure}[t]
\centering
\includegraphics[width=\textwidth]{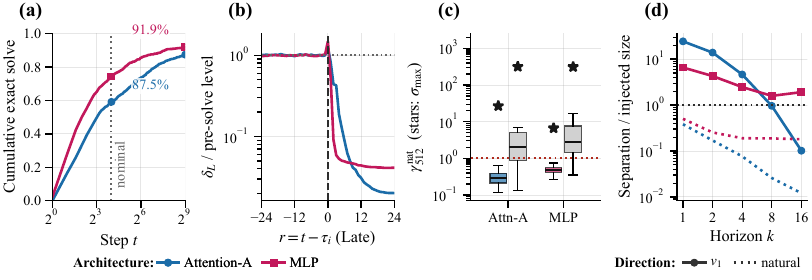}
\caption{\textbf{The completion geometry reproduces across architectures.}
Attention-A in blue, MLP in red. (a)~Cumulative exact solve with extended recurrence, with the nominal horizon marked. (b)~\Late-solver $\delta_L$ on the solve-aligned axis, each model normalized by its own pre-solve level. (c)~Natural-direction gain at step 512 for completed (colored) and
\Persistent (grey) states, with the red line at $\gamma^{\rm nat}=1$ and stars marking median $\sigma_{\max}$ at the same states ($\sigma_{\max}>1$ at every probed state in both models). (d)~Norm-matched perturbations at completed fixed-step states along $v_1$ (solid markers) and along the natural direction (dotted). Both models expand $v_1$ and contract the natural direction, but only Attention-A absorbs the $v_1$ perturbation within 16 steps.}
\label{fig:cross}
\end{figure}

\paragraph{The completion dynamics replicate.} The MLP model is a stronger solver at the nominal horizon, with $74.4\%$ exact solve at step 16 compared with $59.2\%$ for Attention-A. Yet many of its remaining failures are also unfinished computations. By step 512, exact solve reaches $91.9\%$, and 175 of the 256 puzzles unresolved at step 16 are solved later (Figure~\ref{fig:cross}a). The same solve-aligned settling pattern also appears. \Late solvers remain highly mobile at the nominal horizon. Their latent motion drops sharply around their own solution time, and they then enter the same low-motion regime as \Early solvers (Figure~\ref{fig:cross}b). The solving step is again the largest $\zH$ update of the preceding trajectory for almost every \Late solver. Full statistics are given in Appendix~\ref{app:mlp}. 
\paragraph{Trajectory-conditioned stability also replicates.} At step 512, the median natural-direction gain is about $0.47$ for both completed MLP groups, but $2.82$ for \Persistent trajectories. At the same time, $\sigma_{\max}>1$ at every probed MLP state, just as in Attention-A. Completed computations are stable along their own direction of travel,  while the recurrent map still contains expanding directions (Figure~\ref{fig:cross}c). This coexistence occurs at $60.4\%$ of MLP states, close to the $62.1\%$ observed in Attention-A.

The scale of the worst-case expansion does differ. At completed states, $\sigma_{\max}$ is about $6.4$-$6.7$ in the MLP model, compared with about $27$ in Attention-A. Persistent trajectories, by contrast, have similar values in the two models. The qualitative geometry therefore replicates, while its magnitude does not.
\paragraph{The crucial difference lies in the expanding direction.}  In both models, a perturbation along $v_1$ expands much more strongly than one along the natural direction. The difference appears in later recurrent steps. For the 99 completed fixed-step Attention-A states, the $v_1$ perturbation is strongly amplified initially but falls to $0.10$ of its injected magnitude by $k=16$. In the MLP model, it remains at $1.91$ times its injected size (Figure~\ref{fig:cross}d). To determine whether this amplification ultimately decays or persists, the perturbation must be followed beyond the current horizon of $k=16$. Transport shows similar differences: the overlap between the amplified output at one step and the most expanding input at the next is $T_1=1.83\times10^{-3}$ in the MLP model, compared with $4.93\times10^{-5}$ in Attention-A. Attention-A rotates the expanding direction away after completion, whereas the MLP model carries more of it into the next step. Detailed singular-vector statistics and numerical checks are reported in Appendix~\ref{app:mlp}.

\subsection{Replication across tasks and checkpoints}
\label{sec:perturbations}

Our main claims use only Attention-A and MLP, which were analyzed in full detail. As a checkpoint replication, we also run the complete 16-to-512 extended recurrence analysis on the Attention-B \citep{preeti2026checkpoint} checkpoint. Attention-B shows the same trajectory-conditioned anisotropic-stability pattern: its exact-solve rate rises from $0.521$ at step 16 to $0.726$ at step 512, natural-direction gain is below one in the completed groups, and expanding directions remain in the state space. All three extended-recurrence analyses are compared in Appendix~\ref{app:architecture_summary}, Table~\ref{tab:cross}.

We also tested Easy Sudoku and Maze-Hard (Appendices~\ref{app:easy} and~\ref{app:maze}). On Easy Sudoku, computation typically finishes within the first few steps, and the trajectories resemble \Early solvers. The same pattern appears in Maze-Hard with \(30\times30\) grids, where most trajectories settle within the first few steps and again resemble \Early solvers. However, in Maze-Hard, many exact-match failures are valid alternative paths.

Finally, a constraint-aware decoder recovers many puzzles missed by simple argmax decoding (Appendix~\ref{app:decoder}). The TRM decoder independently fills each blank with the highest-scoring digit. To test whether useful alternatives remain in the logits, we evaluate the constraint-aware decoder on Attention-B. For each blank, it keeps the top-$k$ predicted digits and uses backtracking to search only assignments that satisfy the Sudoku row, column, box, and clue constraints. The number of solved puzzles barely changes at $k=1,2,3$, but rises by $22.3$ percentage points at $k=4$. 


\section{Discussion}
Our results show that failure at a fixed inference depth is not equivalent to computational failure. In both models, many puzzles that remain unsolved at step 16 are completed when the same recurrent computation is continued. The latent dynamics also change around completion: state updates decrease sharply, exact solutions typically persist, and the natural trajectory direction becomes locally contracting even though the full Jacobian has strongly expansive directions. Completion, therefore, does not correspond to uniform contraction of the recurrent map. Instead, contraction emerges along the direction followed by the ongoing computation, while unstable directions remain available in the state space.
This qualitative picture is shared by the attention and MLP-based models. Unfinished computations remain dynamically active, whereas completed computations settle, and trajectory-aligned contraction coexists with strong off-trajectory expansion. The clearest difference between the two architectures lies in the fate of perturbations initialized along the locally most expansive direction. In the attention model, this perturbation is rapidly rotated away from the expanding direction and subsequently attenuated, whereas in the MLP model, a larger component persists over later recurrent steps. Thus, a similar trajectory-conditioned stability can coexist with different local perturbation geometries. 

\section{Limitations}
Our results have three main limitations. First, step 512 is still finite, so we cannot determine the asymptotic fate of the \Persistent trajectories. Second, the full Hard-Sudoku analysis covers two attention checkpoints and one MLP checkpoint.  The Maze-Hard provides a separate nominal-horizon task control, but broader claims require more checkpoints, architectures, and tasks. Finally, the Jacobian analysis is local to visited states and uses a subset of the full evaluation set. The most expensive perturbation measurements use smaller subsets.

\section{Conclusion}
A fixed inference budget can obscure the distinction between failed and unfinished computation. In the recursive models studied here, many nominal-horizon failures are solved when the same computation is allowed to continue, and completion is accompanied by a transition to a low-motion, trajectory-stable regime.

This suggests that completion is not the same as uniform contraction of the recurrent map. Instead, stability is tied to the particular path followed by the computation. Looking at these dynamics may help us understand when a recurrent model is still computing, when it has effectively settled, and how it uses its latent state during inference.

\section*{Reproducibility Statement}

We provide the information needed to reproduce the main experimental results in Appendix ~\ref{app:reproducibility} and in the supplementary code. These materials document the datasets, checkpoints, model, sampling, perturbation protocols, control points, and numerical validation used for the Attention-A, Attention-B, and MLP experiments. Statistical procedures are documented in Appendix~\ref{app:stats}. All reported statistics and figures can be traced back to puzzle-level outputs. The supplementary code contains an analysis pipeline that regenerates them from those output archives without requiring model inference. The GPU procedure for regenerating the extended-recurrence results from the original checkpoints is also discussed in Appendix ~\ref{app:reproducibility}.

\section*{AI Use Statement}
Generative AI tools, including OpenAI Codex and Anthropic Claude, were used during several stages of this work. They assisted in implementing experimental and data-analysis code, generating plotting code, refining aspects of the experimental and statistical methodology (including the use and presentation of Wilson intervals, interquartile ranges, and confidence intervals), interpreting numerical results, and improving the clarity and presentation of the manuscript. AI tools also suggested additional diagnostic visualizations, including the analyses reported in Fig.~\ref{fig:stability}c and Fig.~\ref{fig:aligned}c and the correlation between Sudoku rating and solve steps in Appendix ~\ref{app:difficulty}. The research direction, scientific questions, and final methodological and interpretive decisions were determined by the author. All AI-generated or AI-assisted code used for the reported experiments was reviewed and tested, and all reported numerical results and figures were obtained from executed computational experiments. All AI-assisted analyses were reviewed by the authors, who take full responsibility for the correctness of the results, claims, code, and final content of the paper.

\bibliographystyle{iclr2027_conference}
\bibliography{references}

\appendix

\section{Reproducibility and implementation details}
\label{app:reproducibility}
The extended-recurrence results use three complete analyses. \textbf{Attention-A} uses \texttt{step\_21700} and took 91.5 minutes; \textbf{MLP} uses \texttt{step\_16275} and took 82.1 minutes; and \textbf{Attention-B} uses \texttt{step\_65100} and took 88.3 minutes. All three use the same \texttt{hard\_sudoku.csv} data file and \texttt{trm.py} model source, PyTorch 2.11.0+cu128 on one CUDA device. The run uses global seed 0, the joint continuous carry $\zH\oplus\zL$, a nominal horizon of 16, and a diagnostic horizon of 512. Attention-A and Attention-B use PyTorch's mathematical scaled-dot-product attention backend, explicitly pinned as \texttt{SDPBackend.MATH} to prevent hardware-dependent backend selection. The forward recurrence uses bfloat16, while saved latent states and Jacobian, JVP, and VJP calculations use float32. MLP contains no attention operation. The analysis configurations are identical apart from the model and checkpoint paths. For each analysis, we save the full experimental setup in its own \texttt{run\_config.json} and \texttt{run\_manifest.json}. This includes the group definitions, the 150-puzzle probe subset (50 per group), the 36-puzzle expensive subset (12 per group), and all perturbation settings. Jacobian probes use a relative perturbation of $\varepsilon=10^{-4}$, while noise experiments use $\sigma\in\{0.02,0.1\}$ with seeds $\{0,1,2\}$. We also record the use of 20{,}000 bootstrap replicates and five-fold cross-validation. Numerical checks, including the 16-step horizon and the singular-triplet ($\sigma_1,v_1,u_1$) identities, are in \texttt{manifests/numerical\_checks.json} and \texttt{singular\_geometry/table10b\_numerical\_checks.csv} within each archive. Every number in this paper can be traced back to a puzzle-level CSV, with the mapping given in \texttt{ICLR\_RESULT\_INVENTORY.md} for Attention-A and \texttt{MLP\_RESULT\_INVENTORY.md} for MLP. A single CPU, \texttt{python analysis/run\_all.py}, recomputes every reported quantity from those puzzle-level files and regenerates all figures in about a minute, with no model inference. \texttt{REPRODUCE.md} gives the environment, the pinned package versions, and the expected outputs, and \texttt{requirements.txt} pins the analysis environment. Regenerating the results from the checkpoints needs a separate GPU path, which is also documented there. Attention-B follows a similar path. Its entries in Table~\ref{tab:cross} (Appendix~\ref{app:architecture_summary}) come from the extended run with the math SDPA backend, which solves 521 puzzles at step 16. Appendices~\ref{app:snapshots} and~\ref{app:decoder} use an earlier nominal-horizon evaluation of the same checkpoint, which solves 531. Easy Sudoku (Appendix~\ref{app:easy}), Maze-Hard (Appendix~\ref{app:maze}), and the constraint decoder (Appendix~\ref{app:decoder}) come from separate 16-step evaluations.

\section{Statistical procedures}
\label{app:stats}

\paragraph{Statistical procedures.}
We report binary proportions with Wilson 95\% intervals \citep{wilson1927}. Continuous per-puzzle quantities use a percentile bootstrap \citep{efron1979bootstrap} over puzzles with 20{,}000 replicates. Every puzzle is an independent sampling unit.  
Repeated recurrent steps, probe directions, perturbation horizons, and noise seeds are treated as repeated measurements of the same puzzle rather than as independent samples. For solve-aligned analyses, we preserve the pairing between the pre- and post-solve windows by resampling whole puzzles. Jacobian analyses use a fixed subset of 150 puzzles, with 50 puzzles from each completion group. The more expensive direct-perturbation and transport analysis uses 12 puzzles per group. Perturbations injected at step 512 are followed for up to 16 further steps, beyond the 512-step run. Excluding these states moves the Attention-A median $v_1$ separation at $k=16$ from $0.111$ to $0.115$ over all completed states and from $0.102$ to $0.108$ over fixed-step states, and leaves the MLP value of $1.91$ unchanged.


\paragraph{Numerical precision.}
The standard forward pass uses bfloat16, matching the released model checkpoint. For the Jacobian analysis, we upcast the recurrent computation to float32, disable TF32, and use the mathematical SDPA attention backend. In a matched float32 control, the differentiable recurrence exactly reproduces the reference float32 forward pass, with a maximum absolute difference of $0$. 


\paragraph{Terminology.}
We use several dynamical terms in a limited sense. ``Settling'' means that the observed relative state updates become small. It does not imply that the map is contractive. ``Directional gain'' is the one-step Jacobian gain along a specified direction. ``Finite-horizon growth'' describes amplification over a fixed number of future recurrent steps.  ``Completion'' means the first exact decode. A small fraction of trajectories later revert, as quantified in \S\ref{sec:settling}. A ``block'' probe is restricted to $\zL$ or to $\zH$.

\section{TRM Setup}
\label{app:setup}


The attention-based TRM configuration is summarized in Table~\ref{tab:trm-config}. \citet{jolicoeurmartineau2025trm} gives the complete architecture and pseudocode (Figures 1 and 3 of that paper). 

\begin{table}[t]
\centering
\caption{Configuration of the attention-based TRM checkpoints.}
\label{tab:trm-config}
\begin{tabular}{ll}
\toprule
\textbf{Configuration} & \textbf{Value} \\
\midrule
\texttt{H\_cycles} & 3 \\
\texttt{L\_cycles} & 6 \\
\texttt{L\_layers} & 2 \\
\texttt{H\_layers} & 0 \\
Hidden size & 512 \\
Attention heads & 8 \\
MLP expansion factor & 4 \\
Position encoding & Rotary \\
\texttt{puzzle\_emb\_len} & 16 \\
\texttt{puzzle\_emb\_ndim} & 512 \\
\texttt{halt\_max\_steps} & 16 \\
\texttt{no\_ACT\_continue} & \texttt{true} \\
Forward dtype & \texttt{bfloat16} \\
Loss & StableMax cross-entropy  via \texttt{ACTLossHead} \\
Number of parameters & $6{,}829{,}570$ \\
\bottomrule
\end{tabular}
\end{table}

The loss in the above table is StableMax cross entropy loss \citep{prieto2025stablemax}. The input sequence contains 81 grid-cell tokens prefixed by 16 puzzle-embedding tokens, giving 97 token positions. The recurrent carry contains two latent states, $z_H$ and $z_L$, each of shape $97\times512$. Hence the joint continuous carry has dimension
\[
\dim(z_H,z_L)
=
2\times97\times512
=
99{,}328.
\]

In evaluation mode, ACT halts at the final permitted step and resets the carry on the subsequent call. Consequently, obtaining an uninterrupted $N$-step trajectory requires setting the halt cap to $N$. For the long-horizon evaluations in this work, we therefore set \texttt{halt\_max\_steps}=512; see \S\ref{sec:setup}.

\section{Extended-recurrence results}
\label{app:extended}

\subsection*{Solve rate}

The cumulative solve rate is reported in \S\ref{sec:extended}. We distinguish it from the instantaneous solve rate. A puzzle enters the cumulative count as soon as it is solved exactly for the first time. The instantaneous count instead asks whether it is solved at that particular step. These two counts are identical at all reported checkpoints except step 32. By then, 660 puzzles had been solved at least once, while 659 were correct at step 32. One puzzle had briefly reverted after being solved. Since our completion groups are defined by the first exact solve, all group assignments use the cumulative count.

\subsection*{Puzzle difficulty by completion group}
\label{app:difficulty}
\textbf{Datasets.} Hard Sudoku: 1{,}000 puzzles from the test split of Sudoku-Extreme \citep{wang2025hrm} (\url{https://huggingface.co/datasets/sapientinc/sudoku-extreme}), all from its \texttt{puzzles4\_forum\_hardest\_1905} source, which is the \texttt{forum\_hardest\_1905} list of the \texttt{tdoku} benchmark suite \citep{tdoku}. Sudoku-Extreme permutes every puzzle by row, column, box, and digit and keeps its train and test splits mathematically inequivalent.
The dataset carries a per-puzzle \texttt{rating} defined by Sudoku-Extreme as the number of backtracks the \texttt{tdoku} solver \citep{tdoku} needs to solve the puzzle. Table~\ref{tab:difficulty} reports the ratings by completion group.

\begin{table}[h]
\centering\small
\caption{Puzzle difficulty by completion group on the dataset's own rating scale. Difference rows report unpaired median differences relative to \Early solvers, with 20{,}000 bootstrap replicates over puzzles in each group.}

\label{tab:difficulty}
\begin{tabular}{@{}lrrrrrrrl@{}}
\toprule
Group & $n$ & min & Q1 & median & Q3 & max & mean & median 95\% CI \\
\midrule
\Early solver          & 592 & 2 & 14.0 & 27 & 42 & 140 & 31.23 & [25.0,\,28.5] \\
\Late solver           & 283 & 2 & 11.5 & 24 & 39 & 110 & 27.63 & [20.0,\,27.0] \\
\Persistent unsolved   & 125 & 3 & 13.0 & 20 & 28 & 127 & 24.05 & [18.0,\,23.0] \\
All puzzles            & 1000 & 2 & 13.0 & 25 & 40 & 140 & 29.31 & [23.0,\,26.0] \\
\midrule
\multicolumn{9}{@{}l}{\textit{Unpaired median difference against \Early}}\\
\Late $-$ \Early        & 283 & & & $-3$ & & & & [$-7.0$,\,$1.0$] \\
\Persistent $-$ \Early  & 125 & & & $-7$ & & & & [$-9.5$,\,$-3.0$] \\
\bottomrule
\end{tabular}
\end{table}

The puzzles the model never solves are rated \emph{easier} than the ones it solves inside the nominal budget. Two rank statistics say the same thing. Among the 875 eventual solvers, Spearman correlation between rating and first solve time is $\rho=-0.097$ ($p=0.004$): higher-rated puzzles are solved marginally \emph{earlier}. Across all 1{,}000
puzzles, the correlation between rating and ever solving is $\rho=+0.095$ ($p=0.003$). Binned by rating quintile, the \Persistent fraction runs $0.112,\,0.200,\,0.167,\,0.066,\,0.075$ from the lowest to the highest quintile, with no monotone trend. Difficulty therefore does not explain the completion groups. 

\section{Jacobian implementation}
\label{app:jacobian}
Our Jacobian analysis requires differentiating through one outer recurrence of the model. We therefore use a differentiable version of the recurrence in which the \texttt{detach} and \texttt{no grad} boundaries are removed. We verified that this change does not alter the forward computation. After converting the relevant operations to float32, the differentiable and original implementations agree exactly at the level of the model state.

We estimate the leading singular value of the Jacobian, $\sigma_1$, using matrix-free power iteration on $J^\top J$. This avoids constructing the full Jacobian, which would be prohibitively large for our state space. Jacobian-vector and vector-Jacobian products are computed with \texttt{jvp} and \texttt{vjp}, respectively \citep{baydin2018autodiff}. We use 10 power iterations with a convergence tolerance of $10^{-3}$. Once the leading right singular vector $v_1$ is obtained, the corresponding left singular vector is defined as
\begin{equation}
u_1=\frac{Jv_1}{\sigma_1}.
\end{equation}

Because the quantities of interest are sensitive to small numerical errors, we perform these calculations in float32 and disable TF32. We also use the mathematical SDPA backend for attention rather than the fused Flash, memory-efficient, or cuDNN SDPA backends.


We carried out numerical checks on the estimated singular triplets. Across 1,100 probe entries, the singular vectors remain normalized to numerical precision, and $98.6\%$ of the triplets satisfy the scalar convergence criterion based on relative changes in $\sigma_1$. The relation $Jv_1=\sigma_1u_1$ holds by construction. The adjoint norm ratio $\|J^\top u_1\|_2/\sigma_1$ remains within $1.3\%$ of one across these entries.


Taken together, these checks suggest that the matrix-free procedure is sufficiently stable for the Jacobian analyses reported in the paper.

\section{Latent block perturbations and finite-horizon growth}
\label{app:blocks}

\begin{figure}[h]
\centering
\includegraphics[width=\textwidth]{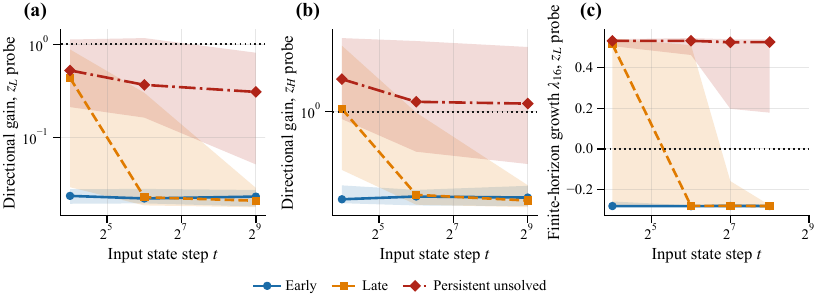}
\caption{(a,b)~Median one-step directional gain for probe directions supported on $\zL$ and $\zH$, shown by group and input-state step. (c)~Finite-horizon growth $\lambda_{16}$ following a perturbation to $\zL$. Shaded regions show
interquartile ranges.}
\label{fig:blocks}
\end{figure}



To probe stability beyond one step, we measure the finite-horizon
separation of a clean and a perturbed trajectory. For each probed state
$s_t$, we draw a isotropic unit direction $v_L$ supported
entirely in $\zL$ and set
\[
v=(0,v_L), \qquad
\widetilde{s}_t=s_t+\varepsilon_t v,
\qquad
\varepsilon_t=10^{-4}\|s_t\|_2.
\]
After evolving both states under the same recurrent map for $k$ steps, let
\[
\Delta s_{t+k}^{(v)}
    =\widetilde{s}_{t+k}-s_{t+k}.
\]
We define
\[
\lambda_k(v)
  =\frac{1}{k}\log
    \left(
      \frac{\|\Delta s_{t+k}^{(v)}\|_2}{\varepsilon_t}
    \right).
\]
Negative values mean that the separation at horizon $k$ is smaller than
the injected perturbation, while positive values mean that it is larger.
This is a finite-horizon, trajectory-conditioned measure of perturbation
growth, not an asymptotic Lyapunov exponent.\\

Latent block-supported probes show how the response differs between the two latent components (Figure ~\ref{fig:blocks} a,b). At step 16, the median gain for \Early trajectories is below one in both blocks. The separation from \Persistent trajectories is more pronounced in $\zH$, where the median gains are $0.190$ for \Early and $1.851$ for \Persistent trajectories.

By step 512, the \Early and \Late groups have similar responses in both latent blocks. \Persistent trajectories retain larger gains in $\zH$ with a median greater than one.

Finite-horizon growth shows the same distinction (Figure~\ref{fig:blocks} c). For $\zL$ perturbations introduced at step 256, the median growth over the next 16 steps is $\lambda_{16} = -0.282$ in both the \Early and \Late groups and $0.526$ in the \Persistent group.

These quantities describe growth along the observed recurrent trajectory over a fixed number of future steps. They should therefore be interpreted as finite-horizon, trajectory-conditioned rates rather than asymptotic Lyapunov exponents \citep{vogt2022lyapunov,engelken2023lyapunov}.

\section{Full MLP model cross-architecture replication}
\label{app:mlp}

In this appendix, we study the MLP experiments in full detail. Relevant quantities are defined in \S\ref{sec:setup} and computed on the same task and data.


\subsection{Checkpoint and configuration for MLP}
\label{app:mlp_setup}
The configuration of the MLP model used in our experiments is summarized in Table~\ref{tab:mlp-config}. Except for the token-mixing mechanism and positional encoding, its architectural and evaluation settings are identical
to those of Attention-A. Both runs use the same puzzles file and the same model source.

\begin{table}[t]
\centering
\caption{Configuration of the MLP model used in the experiments.}
\label{tab:mlp-config}
\begin{tabular}{ll}
\toprule
\textbf{Configuration} & \textbf{Value} \\
\midrule
Checkpoint & \texttt{step\_16275} \\
Number of parameters & $5{,}030{,}402$ \\
Attention-A parameters & $6{,}829{,}570$ \\
Token mixing & Sequence-axis MLP \\
\texttt{mlp\_t} & \texttt{true} \\
Position encoding & None \\
\texttt{H\_cycles} & 3 \\
\texttt{L\_cycles} & 6 \\
\texttt{L\_layers} & 2 \\
Hidden size & 512 \\
MLP expansion factor & 4 \\
Forward dtype & \texttt{bfloat16} \\
Loss & StableMax cross-entropy \\
Nominal recurrence length & 16 steps \\
\bottomrule
\end{tabular}
\end{table}



The MLP replaces attention-based token mixing with a sequence-axis MLP and omits rotary positional encoding. The remaining architectural settings match Attention-A as shown in Table~\ref{tab:mlp-config}.

The analysis uses global seed 0 on a CUDA device. Over all 512,000 recorded step observations, no non-finite values occur, and both latent-state norms remain bounded. After conversion to float32, the differentiable recurrence also matches the original forward computation exactly. 

\subsection{Extended recurrence and completion groups}
\label{app:mlp_extended}
\begin{figure}[h]
\centering
\includegraphics[width=\textwidth]{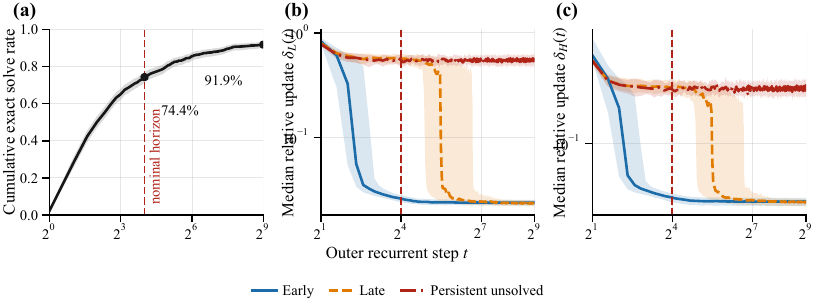}
\caption{MLP extended recurrence. (a)~Cumulative exact solve with Wilson bands. (b,c)~Median relative latent update by completion group with interquartile bands.}
\label{fig:mlp_extended}
\end{figure}

Extending the recurrence from 16 to 512 steps raises the number of puzzles solved at least once from 744 to 919 (Figure~\ref{fig:mlp_extended}a). Moreover, blank-cell accuracy improves while StableMax loss decreases.

Of the 1,000 puzzles, $744$ belong to the \Early group
\cip{71.6\%}{77.0\%}, $175$ to the \Late group
\cip{15.3\%}{20.0\%}, and $81$ remain \Persistent
\cip{6.6\%}{10.0\%}. Among the puzzles that eventually solve, the median first solve time is $3$ steps for \Early trajectories (IQR 2-6) and $43$ steps for
\Late trajectories (IQR 28-100).

All $919$ MLP puzzles that ever reach an exact solution keep it: $P(\text{stays solved})=1.000$ \cip{0.9958}{1.0}, with zero reversions, against $859/875 = 0.982$ in Attention-A. 

The latent dynamics yield the same ordering as in Attention-A (Figure~\ref{fig:mlp_extended}b,c). \Early trajectories already have small updates at step 16 while \Late and \Persistent trajectories remain more active. At step 512, the \Late group reaches a similar low-motion regime to the \Early group, whereas \Persistent trajectories continue to change substantially. 

The difficulty rating is only weakly associated with these outcomes. Among the 919 eventual solvers, higher ratings are associated with slightly earlier first solutions $(\rho, p) = (-0.105, 0.0014)$. The association with whether a puzzle ever solves is small $(\rho=0.025, p =0.43)$. These results do not support a simple interpretation in which later or unsuccessful trajectories are harder puzzles on this rating scale.

\subsection{Solve-aligned settling and the solving step}
\label{app:mlp_aligned}

\begin{figure}[h]
\centering
\includegraphics[width=\textwidth]{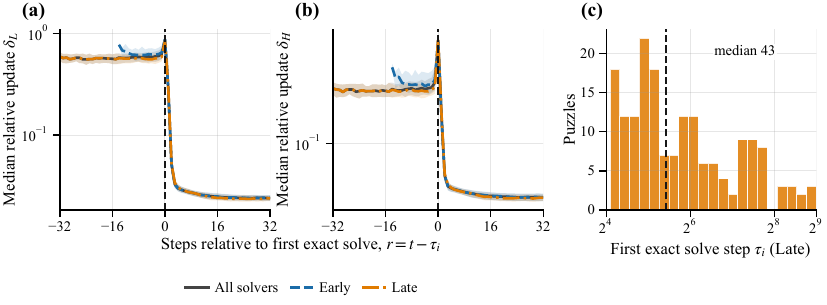}
\caption{MLP solve-aligned dynamics. (a,b)~Relative updates along the solve-aligned axis. (c)~Distribution of first solve times for \Late solvers.}
\label{fig:mlp_aligned}
\end{figure}

The MLP shows a sharp reduction in latent motion around completion (Figure~\ref{fig:mlp_aligned}a,b). Among the 175 \Late solvers, the median paired change in window-mean $\delta_L$ between the pre-solve window $[-8, -1]$ and post-solve window $[0, 7]$ is $-0.399$ \ci{-0.408}{-0.391}. The corresponding change in $\delta_H$ is $-0.138$ \ci{-0.147}{-0.132} compared with $-0.029$ in Attention-A.

The earlier comparison between $[-16, -9]$ and $[-8, -1]$ shows little change in $\delta_L$: 0.006 \ci{-0.007}{0.014}. The reduction is concentrated around completion rather than spread across the preceding windows. After solving, \Late trajectories approach a low-motion regime similar to \Early trajectories. 

The solving step itself contains a pronounced state update. The paired increase in $\delta_H$ from $r=-1$ to $r=0$ is 0.508 \ci{0.498}{0.523}. For $99.4\%$ of \Late solvers, the $\zH$ update at the solving step is larger than at every earlier step. The solving step $\zL$ update is the largest observed up to that step for 74.3\% of \Late MLP trajectories, compared with 35.3\% in Attention-A.


\subsection{Fixed-time Jacobian, natural-direction gain, and \texorpdfstring{$\sigma_{\max}$}{sigma}}
\label{app:mlp_jacobian}

\begin{figure}[h]
\centering
\includegraphics[width=\textwidth]{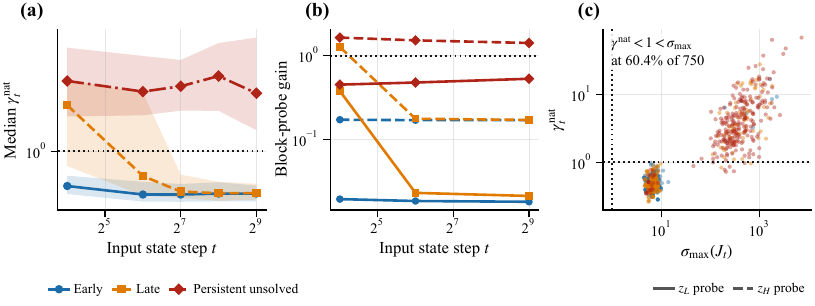}
\caption{MLP local stability at fixed recurrent steps. (a)~Median natural-direction gain by group. (b)~Probe gains supported on $\zL$ (solid) and $\zH$ (dashed). (c)~Natural-direction gain $\gamma^{\rm nat}_t$ versus
$\sigma_{\max}(J_t)$ for all probed states.}
\label{fig:mlp_jacobian}
\end{figure}

The natural-direction gain already separates the three groups at the nominal 16-step horizon (Figure~\ref{fig:mlp_jacobian}). Median $\gamma^{\rm nat}$ is $0.538$ \ci{0.494}{0.607} for \Early, $2.271$ \ci{1.347}{2.858} for \Late, and $3.505$ \ci{2.450}{4.635} for \Persistent trajectories. Accordingly, $92\%$ of \Early states have gain below one, compared with $32\%$ of \Late states and $12\%$ of \Persistent states.




By step 512, median natural-direction gains are similar in the \Early and \Late groups ($0.470$ and $0.473$) while remaining above one in the \Persistent group ($2.816$). Nevertheless, $\sigma_{\max} > 1$ at all 750 probed states. Small gains along the natural direction coexist with expanding directions in the surrounding state space (Figure~\ref{fig:mlp_jacobian}c).

Latent block-supported probes show the same group ordering. The \Early and \Late groups have smaller median gains than the \Persistent group in both latent components (Figure~\ref{fig:mlp_jacobian}b). 

The finite-horizon measurements extend this comparison beyond one step. For $\zL$ perturbations introduced at step 256, growth over the following 16 steps is negative in 98\% of \Early states, 90\% of \Late states and 12\% of \Persistent states.






\subsection{Leading singular-vector geometry}
\label{app:mlp_geometry}

\begin{figure}[h]
\centering
\includegraphics[width=\textwidth]{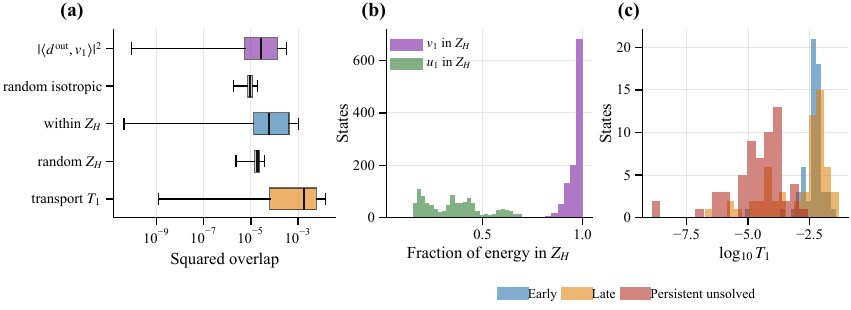}
\caption{MLP leading singular-vector geometry. (a)~Squared overlaps and one-step transport for the fixed-time states, together with random baselines. The ``random isotropic'' is the overlap of $v_1$ with a random unit direction drawn uniformly from the full hidden-state space, and ``random \(\zH\)'' uses a random unit direction drawn from the \(\zH\) subspace only. Transport is evaluated on the 180-state subset. (b)~Fraction of the leading right and left singular vectors supported on $\zL$ and $\zH$. (c)~Distribution of $\log_{10}T_1$ across completion groups.}
\label{fig:mlp_geometry}
\end{figure}


The natural update has a small overlap with $v_1$, but it exceeds the random isotropic baseline (Figure~\ref{fig:mlp_geometry}a). The median squared overlap is $2.55\times 10^{-5}$ compared with $9.46 \times 10^{-6}$ for the random baseline. The paired difference is $1.61 \times 10^{-5}$ \ci{1.33\times 10^{-5}}{2.19\times 10^{-5}}. The $\zH$-restricted comparison shows the same pattern. 

In both architectures, the leading input direction $v_1$ lies almost entirely in $\zH$, while its amplified output $u_1$ lies mainly in $\zL$ (Figure~\ref{fig:mlp_geometry}b). The $\zL$ component contains 65\% of the output-direction energy in the MLP compared with 85.3\% in Attention-A. The leading response is less concentrated in $\zL$ for the MLP.

The leading direction is transported more strongly between successive steps in the MLP (Figure~\ref{fig:mlp_geometry}c). Over the 180-state subset, median $T_1$ is $1.83 \times 10^{-3}$ \ci{5.7 \times 10^{-4}}{4.6 \times 10^{-3}}, compared with $4.93 \times 10^{-5}$ in Attention-A.

\subsection{Direct \texorpdfstring{$v_1$}{v1} versus natural-direction perturbations}
\label{app:mlp_perturbation}

\begin{figure}[h]
\centering
\includegraphics[width=0.72\textwidth]{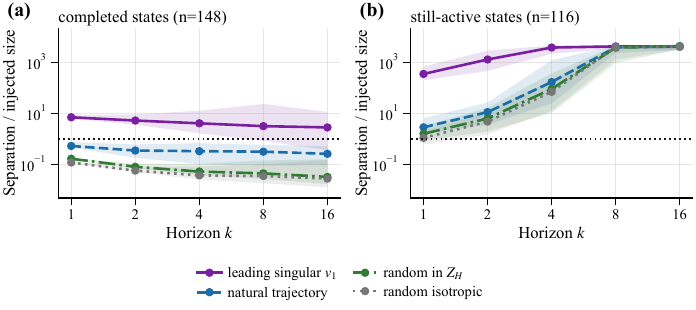}
\caption{MLP response to norm-matched perturbations with initial size $10^{-4}$ of the state norm. Results are separated according to whether the unperturbed computation has already completed. Panel (a) pools the 100 fixed-step and 48 solve-aligned completed states (the text quotes fixed-step medians), and panel (b) shows the 116 still-active states. Shaded regions show interquartile ranges.}
\label{fig:mlp_perturbation}
\end{figure}

At the 100 completed fixed-step states, perturbations along $v_1$ initially expand, with a median one-step amplification of $6.58$ \ci{6.43}{6.96} (Figure~\ref{fig:mlp_perturbation}a). After 16 steps, the median separation is $1.91$ \ci{0.87}{2.74} times the injected size. Although the point estimate remains greater than one, its confidence interval spans one, and $42\%$ of states have separations less than the injected size. Recovery also varies between completion groups: the corresponding medians are $0.91$ for \Early states and $3.10$ for \Late states.

Natural-direction perturbations contract more consistently. After 16 steps, their median separation is $0.181$ \ci{0.113}{0.234} times the injected size, with $93\%$ of states below one. Both random controls contract more strongly still.

At the 116 still-active probe states, all four perturbation directions produce separations of approximately $4 \times 10^3$ times the injected size by step 16. Given the initial scale of $10^{-4}$ of the state norm, these separations are of the order of the state norm (Figure~\ref{fig:mlp_perturbation}b). 

These results distinguish initial amplification from subsequent recovery. The less consistent recovery of $v_1$ perturbations in the MLP is compatible with its larger measured direction transport.

\subsection{Cross-checkpoint and architecture summary}
\label{app:architecture_summary}
\begin{table}[h]
\centering\small
\setlength{\tabcolsep}{4pt}
\caption{Cross-checkpoint and cross-architecture comparison. Completion-group quantities use step 512. $\gamma^{\rm nat}$ and $\sigma_{\max}$ are medians over 50 probed states per
group. Perturbation separations are medians over completed fixed-step probed states. A range in a completed entry gives the Early-Late group medians.}
\label{tab:cross}
\begin{tabular}{@{}lrrr@{}}
\toprule
Quantity & Attention-A & MLP & Attention-B \\
\midrule
Step-16 / step-512 exact solve
  & 0.592 / 0.875 & 0.744 / 0.919 & 0.521 / 0.726 \\
Step-16 unresolved with any later exact solve
  & 283/408 = 0.694 & 175/256 = 0.684 & 206/479 = 0.430 \\
Solutions never lost after first solve
  & 859/875 = 0.982 & 919/919 = 1.000 & 719/727 = 0.989 \\
\midrule
$\gamma^{\rm nat}$, completed / \Persistent
  & 0.25-0.31 / 2.09 & 0.47 / 2.82 & 0.20-0.21 / 2.16 \\
$\sigma_{\max}$, completed / \Persistent
  & 26.8-26.9 / 316.9 & 6.4-6.7 / 312.4 & 17.6-19.0 / 283.9 \\
$\Pr[\gamma^{\rm nat}<1<\sigma_{\max}]$
  & 0.621 & 0.604 & 0.624 \\
\midrule
$v_1$ perturbation, $k{=}1$ / $k{=}16$
  & 24.4 / 0.10 & \phantom{0}6.6 / 1.91 & 21.1 / 0.096 \\
Natural perturbation, $k{=}1$ / $k{=}16$
  & 0.38 / 0.012 & \phantom{0}0.50 / 0.18 & 0.23 / 0.0070 \\
Transport $T_1$
  & $4.9\times10^{-5}$ & $1.8\times10^{-3}$ & $1.4\times10^{-4}$ \\
\bottomrule
\end{tabular}
\end{table}

Table~\ref{tab:cross} summarizes the shared findings and differences between checkpoints. Extended recurrence resolves many nominal-budget failures. In Attention-A and the MLP, \Late trajectories settle after solving and approach the low-motion regime of \Early trajectories. Completed groups typically have the natural-direction gains below one even though expanding directions remain in the local Jacobian. 

The checkpoints differ in solve rates, the size of the completion transition, and the response to perturbations. In particular, the MLP has greater transport of the leading expanding direction and less consistent recovery of perturbations along that direction over 16 steps.

\section{Attention-B training snapshots}
\label{app:snapshots}
This section reports the separate nominal-horizon sweep over all 25 released Attention-B training snapshots \citep{preeti2026checkpoint}. The full extended analysis of the terminal checkpoint is reported in Appendix~\ref{app:architecture_summary}, Table~\ref{tab:cross}. The snapshots span training steps 2{,}604 to 65{,}100. Most are spaced by 2{,}604 training steps, except that the third snapshot is at step 6{,}510.

\begin{figure}[!ht]
\centering
\includegraphics[width=\textwidth]{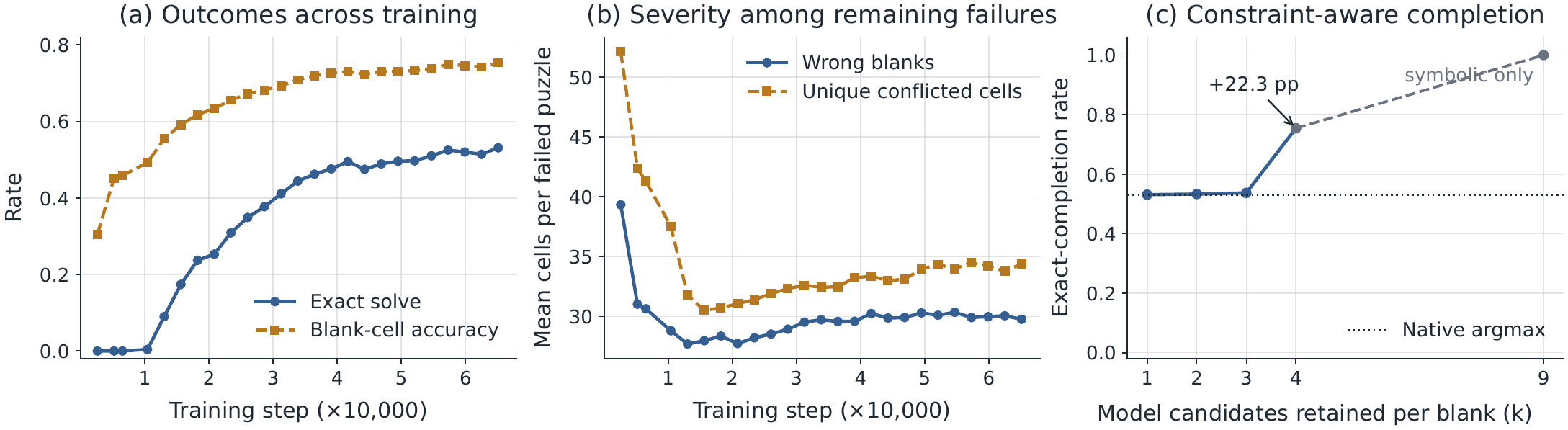}
\caption{Attention-B training and decoder diagnostics. (a) Exact-solve and blank-cell accuracy across 25 correlated
training snapshots. (b) Conflict severity. (c) Exact completion under model-guided Sudoku constraints; the dotted line is native argmax and $k=9$ is a symbolic-only control.}
\label{fig:snapshots}
\end{figure}

Nominal-horizon exact-solve performance improves overall across training, reaching 53.1\% at the final checkpoint, with several temporary declines (Figure~\ref{fig:snapshots} a). Conflict severity among unsolved puzzles initially decreases and later rises slightly (Figure~\ref{fig:snapshots} b), although the set of unsolved puzzles changes between snapshots.

\subsection{Comparison between checkpoints at the nominal horizon}
\label{app:architecture}

The 25-snapshot sweep provides a training-time view of the outcome-associated trends seen in Attention-A. The terminal Attention-B checkpoint also provides an independent full-run replication of the same attention architecture, summarized in Appendix~\ref{app:architecture_summary}.

The three Sudoku checkpoints are as follows. \textbf{Attention-A} (\texttt{step\_21700}) is the checkpoint studied in the main text \citep{sanjin2024checkpoint}. \textbf{Attention-B} (\texttt{step\_65100}) comes from a separate public training run of the same architecture \citep{preeti2026checkpoint}. \textbf{MLP} (\texttt{step\_16275}) replaces attention-based token mixing with an MLP along the sequence axis and uses no positional encoding \citep{sanjin2025checkpointmlp}.

All three models use hidden width 512, three $H$-cycles, six $L$-cycles, two layers in the repeated $L$-level module, and bfloat16 forward computation. The two attention models use eight attention heads and rotary positional encodings \citep{su2021roformer}.

\begin{table}[h]
\centering\small
\caption{Nominal 16-step evaluation on 1,000 hard Sudoku puzzles. Final $\zL$ updates are reported as medians within the solved and failed groups. ``Factor'' is the ratio of the failed-group median to the solved-group median.}
\label{tab:architecture}
\begin{tabular}{@{}lrrrr@{}}
\toprule
Checkpoint & Exact solve & Final $\zL$ update, solved / failed & Factor  \\
\midrule
Attention-A & 0.592 & 0.025 / 0.761 & $\sim$30  \\
Attention-B & 0.531 & 0.069 / 0.708 & 10.2  \\
MLP    & 0.744 & 0.018 / 0.536 & $\sim$30  \\
\bottomrule
\end{tabular}
\end{table}

The MLP achieves the highest nominal solve rate followed by Attention-A and Attention-B. All three checkpoints have larger median final $\zL$ updates on failed puzzles than on solved puzzles (Table~\ref{tab:architecture}). 
\section{Constraint-aware decoding}
\label{app:decoder}

The TRM decoder fills each blank with the highest-scoring digit independently. Related neural Sudoku methods incorporate constraint structure through recurrent message passing \citep{palm2018rrn} or differentiable
satisfiability layers \citep{wang2019satnet}. To test whether useful alternatives remain in the logits, we also evaluate a constrained decoder on Attention-B. For each blank, it keeps the top-$k$ predicted digits and uses backtracking to search only assignments that satisfy the Sudoku row, column, box, and clue constraints. The search is capped at 200{,}000 nodes per puzzle (Table~\ref{tab:decoder}).

\begin{table}[h]
\centering\small
\caption{Constraint decoding for Attention-B on $1{,}000$ puzzles. ``Supported'' means that the correct digit for every blank appears in that blank's top-$k$ predictions. The $k=9$ setting removes model-based pruning and serves only as a symbolic-solver control.
}
\label{tab:decoder}
\begin{tabular}{@{}rrrrr@{}}
\toprule
$k$ & Supported & Exact & Median nodes & 95th pct.\ nodes \\
\midrule
1 & 0.531 & 0.531 & 56 & 59 \\
2 & 0.533 & 0.533 & 56 & 59 \\
3 & 0.537 & 0.537 & 59 & 659 \\
4 & 0.754 & 0.754 & 59 & 9{,}304 \\
9 & 1.000 & 1.000 & 59 & 16{,}049 \\
\bottomrule
\end{tabular}
\end{table}

All 469 failures of the native decoder are complete boards that violate Sudoku constraints. Allowing two or three candidate digits per blank changes the solve rate slightly. With four candidates, the constrained decoder finds 754 exact solutions, recovering 223 cases missed by native argmax decoding (Table ~\ref{tab:decoder}).

These recoveries show that useful alternatives remain in the logits. A correct completion can be found within the top-four candidate sets when Sudoku constraints are enforced by backtracking. The choice of $k=4$ was made using the same 1,000 evaluation puzzles.

\section{Easy Sudoku }
\label{app:easy}

The Easy Sudoku control uses the 100 puzzles in the \texttt{nikoli\_100} set of Sudoku-Bench
\citep{seely2025sudokubench} (\url{https://huggingface.co/datasets/SakanaAI/sudoku-bench-nikoli}). These puzzles contain 46-58 blanks compared with 55-60 in the hard split. 

Both Attention-A and the MLP solve all 100 puzzles within the nominal 16 steps. At the first step, Attention-A solves 21 puzzles and the MLP solves 66. The final latent updates are small in both models, although their magnitudes differ (Table~\ref{tab:easy}). These updates are comparable to those of Hard Sudoku puzzles after they are solved, for both \Early and \Late solvers. 


\begin{table}[h]
\centering\small
\caption{Easy Sudoku control, $n=100$, at the nominal 16-step horizon.
}
\label{tab:easy}
\begin{tabular}{@{}lrr@{}}
\toprule
Quantity & Attention-A & MLP \\
\midrule
Exact-solve rate & 1.000 & 1.000 \\
Solved at step 1 & 21 & 66 \\
Steps to $99\%$ of final mean blank-cell accuracy & 6 & 2 \\
\midrule
$\zL$ final update & 0.017 & 0.024 \\
$\zH$ final update & 0.008 & 0.034 \\
\bottomrule
\end{tabular}
\end{table}



\section{Maze-Hard}
\label{app:maze}

Maze-Hard provides a cross-task control using 1{,}000  $30 \times 30$ mazes from the benchmark of \citet{wang2025hrm}
(\url{https://huggingface.co/datasets/sapientinc/maze-30x30-hard-1k}). Each cell holds one of six symbols. We evaluate an attention checkpoint (\texttt{step\_9765}; \citealp{sanjin2025checkpointmaze}) with three $H$-cycles and four $L$-cycles.

\begin{figure}[h]
\centering
\includegraphics[width=\textwidth]{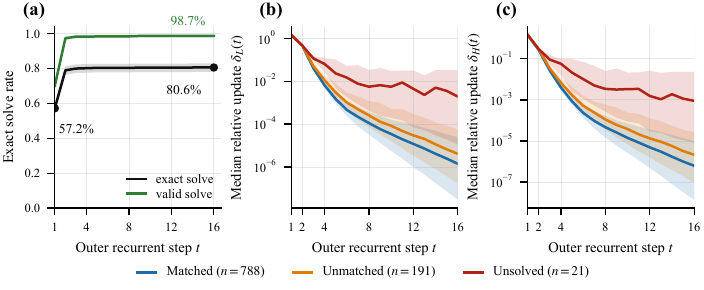}
\caption{Maze-Hard outcomes and latent settling. (a)~Cumulative exact solves
from bfloat16 run, with Wilson bands. Figures (b,c)~Median relative latent updates in float64. Shaded bands show interquartile ranges.}
\label{fig:maze_settling}
\end{figure}

Most solutions that match the ground truth appear early, with 800 mazes matching the ground truth at step 3 (790 at step 2). Over the entire 16-step evaluation, 806 mazes match the ground truth at least once, but only 788 do so at the final step because 18 earlier solutions are subsequently lost. This 16-step evaluation does not determine whether the unresolved cases would solve if recurrence were extended beyond step 16.

Exact match with the ground truth also understates path validity. Of the 212 final non-matches, 191 are valid start-to-goal paths, 75 of them as short as the ground truth, and only 21 are structurally invalid. 


\begin{table}[h]
\centering\small
\caption{Maze-Hard results, $n=1{,}000$, at the nominal 16-step horizon. Intervals are Wilson 95\%.}
\label{tab:maze}
\begin{tabular}{@{}lr@{}}
\toprule
Metric & Value \\
\midrule
Exact-match rate & 0.788 [0.762,\,0.812] \\
Valid-path rate & 0.979 [0.968,\,0.986] \\
Valid but not exact & 191 \\
\quad equal path length to reference & 75 \\
\quad longer (mean $+2.1$ cells) & 116 \\
Structurally invalid paths & 21 \\
Exactly correct earlier, lost by step 16 & 18 \\
\bottomrule
\end{tabular}
\end{table}
\subsection*{Latent settling}
We categorize the puzzle in three groups: \Matched, \Unmatched, and \Unsolved. \Matched means the model solved the puzzle and it matches the ground truth $( n=788)$. \Unmatched means the solution provided by the model is a valid path but doesn't match the ground truth $( n=191)$. \Unsolved means the model failed to provide the structurally valid solution to the puzzle $( n=21)$. 

At this scale, the measurements cannot reliably resolve differences in
settling. We therefore replay the recurrence from the checkpoint's initial
state in float64. We keep the groups and solve times fixed,
so that the control does not also change the outcome definition.
\paragraph{Accuracy:}
Acc. is all-cell accuracy: the fraction of the 900 maze cells whose predicted label matches the reference. It includes walls, open cells, endpoints, and path cells. Many unchanged background cells can make this high even when the route is wrong.
Path F1 measures overlap between predicted and reference path-marked cells:
\begin{equation}
\mathrm{Path\ F1}=\frac{2|P\cap R|}{|P|+|R|},
\end{equation}
where \(P\) and \(R\) are the predicted and reference path-cell sets. Start and goal cells have separate labels and are excluded from these sets.                                                                                                                                                                                                                                                                                                                                                                                                                                                                                                            
The table (\ref{tab:maze_groups}) reports the mean per-puzzle score within each group at step 16, using the BF16 predictions.
Importantly, Path F1 is not path validity. A valid alternative route can score below 1 because it differs from the reference. An invalid route can score highly because most of its cells overlap the reference. That explains why the Unsolved group can have higher accuracy and Path F1 than the Unmatched group. The change in latents $\zL$ and $\zH$ for these categories are also reported in table (\ref{tab:maze_groups}) and plotted in fig. (\ref{fig:maze_settling} b, c). 

\begin{table}[t]
\centering\small
\setlength{\tabcolsep}{6pt}
\caption{Maze outcomes and latent motion at step 16. Groups, mean all-cell accuracy, and mean path F1 come from the  bfloat16 run. Median relative latent updates $\delta_L$ and $\delta_H$ come from the float64 evaluation.}
\label{tab:maze_groups}
\begin{tabular}{@{}lccccc@{}}
\toprule
Group & $n$ & Acc. & Path F1 & $\delta_L$ & $\delta_H$ \\
\midrule
\textsc{Matched} & 788 & 1.000 & 1.000 & $1.4\!\times\!10^{-6}$ & $6.2\!\times\!10^{-7}$ \\
\textsc{Unmatched} & 191 & 0.968 & 0.871 & $4.1\!\times\!10^{-6}$ & $2.1\!\times\!10^{-6}$ \\
\textsc{Unsolved} & 21 & 0.975 & 0.901 & $1.9\!\times\!10^{-3}$ & $9.0\!\times\!10^{-4}$ \\
\bottomrule
\end{tabular}
\end{table}

\begin{figure}[!htb]
\centering
\includegraphics[width=\textwidth]{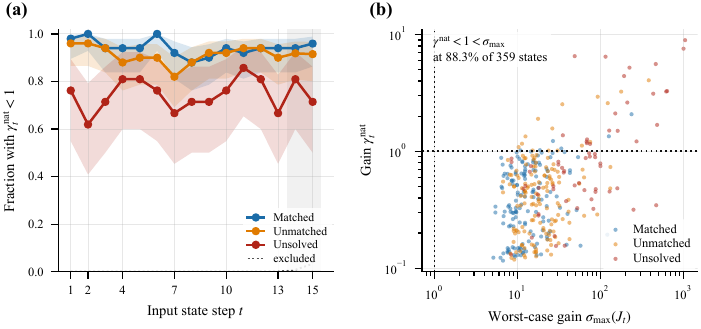}
\caption{Directional contraction and worst-case expansion.
(a)~Fraction of valid directions with $\gnat_t<1$, with Wilson
bands. (b)~States with both a valid directional gain and a singular-value
estimate, at $t=2,7,15$. The inequalities $\gnat_t<1<\sigma_{\max}(J_t)$
hold together at $317/359$ states ($88.3\%$).}
\label{fig:maze_anisotropy}
\end{figure}

\paragraph{Local stability:} Small latent updates alone do not establish local stability. We investigate whether a small displacement along the trajectory is contracted by the next
recurrent step. Let $s_t=(Z_{H,t},Z_{L,t})$ be the joint latent state, with
the puzzle held fixed
\begin{equation}
J_t=DF_\theta(s_t),\qquad
\dout_t=\frac{s_{t+1}-s_t}{\|s_{t+1}-s_t\|_2},\qquad
\gnat_t=\|J_t\dout_t\|_2 .
\end{equation}
 A gain $\gnat_t < 1$ indicates local contraction along that direction.
We compare it with $\sigma_{\max}(J_t)$, the largest gain over all directions.
The latter is estimated by power iteration. For contractions, see table \ref{tab:maze_stability} and fig. \ref{fig:maze_anisotropy} a, b for more information.



\begin{table}[t]
\centering\small
\setlength{\tabcolsep}{4pt}
\caption{Local stability by input-state step. $\gnat$ measures gain along the trajectory. $\sigma_{\max}$ estimates the largest gain over all directions. Entries are medians with IQR ranges. $n_\gamma$ counts valid directions. Each measured $\sigma_{\max}$ entry uses all 50, 50, or 21 states.}
\label{tab:maze_stability}
\begin{tabular}{@{}llrccc@{}}
\toprule
Step & Group & $n_\gamma$ & $\gnat$ & $\Pr[\gnat<1]$ & $\sigma_{\max}$ \\
\midrule
\multirow{3}{*}{2}
 & \textsc{Matched} & 50 & 0.15 {\scriptsize[0.14,\,0.21]} & 1.00 & 12.5 {\scriptsize[9.2,\,19.9]} \\
 & \textsc{Unmatched} & 50 & 0.18 {\scriptsize[0.14,\,0.25]} & 0.96 & 21.5 {\scriptsize[12.7,\,28.9]} \\
 & \textsc{Unsolved} & 21 & 0.51 {\scriptsize[0.32,\,3.12]} & 0.62 & 66.3 {\scriptsize[38.5,\,223.0]} \\
\midrule
\multirow{3}{*}{7}
 & \textsc{Matched} & 50 & 0.53 {\scriptsize[0.30,\,0.76]} & 0.92 & 11.3 {\scriptsize[8.8,\,17.7]} \\
 & \textsc{Unmatched} & 50 & 0.54 {\scriptsize[0.36,\,0.77]} & 0.82 & 16.1 {\scriptsize[11.7,\,25.1]} \\
 & \textsc{Unsolved} & 21 & 0.79 {\scriptsize[0.51,\,1.18]} & 0.67 & 74.3 {\scriptsize[27.0,\,114.9]} \\
\midrule
\multirow{3}{*}{15}
 & \textsc{Matched} & 49 & 0.61 {\scriptsize[0.38,\,0.77]} & 0.96 & 11.7 {\scriptsize[8.9,\,17.5]} \\
 & \textsc{Unmatched} & 47 & 0.59 {\scriptsize[0.43,\,0.76]} & 0.91 & 15.5 {\scriptsize[12.1,\,23.1]} \\
 & \textsc{Unsolved} & 21 & 0.75 {\scriptsize[0.48,\,1.27]} & 0.71 & 68.8 {\scriptsize[27.0,\,99.7]} \\
\bottomrule
\end{tabular}
\end{table}


\vspace{2cm}

\paragraph{A larger gain can accompany smaller motion.}
For \Matched puzzles, median gain rises from $0.15$ at $t=2$ to $0.61$
at $t=15$ (Table~\ref{tab:maze_stability}). Over the same steps, the
full-group median $\delta_L$ falls from $4.4\times10^{-1}$ to
$2.5\times10^{-6}$. There is no contradiction. Gain is a local
amplification factor, whereas the update measures how far the state moves.
These two also uses different populations. The update is a blockwise (either $\zL$ or $\zH$)
quantity over all 788 \Matched puzzles. The gain uses selected joint-state
probes.



\paragraph{Stability is directional, not global.}
The contraction along the trajectory coexists with expansion
in another direction at $317/359$ states with both measurements
($88.3\%$; Figure~\ref{fig:maze_anisotropy}). The Wilson 95\% interval is $84.6\%$--$91.2\%$.

\end{document}